\documentclass[runningheads]{llncs}
   
\usepackage{graphicx}
\usepackage{textcomp}
\usepackage{array,ragged2e}

\def\BibTeX{{\rm B\kern-.05em{\sc i\kern-.025em b}\kern-.08em
    T\kern-.1667em\lower.7ex\hbox{E}\kern-.125emX}}

\usepackage{colortbl}
\usepackage{makecell}
\usepackage{threeparttable}
\usepackage{pgfplots}
\pgfplotsset{compat=1.17}
\usepackage{amsmath}
\usepackage{multicol}
\usepackage{multirow}
\usepackage{caption}
\usepackage{tabto}
\usepackage{wrapfig}
\usepackage{amssymb}
\usepackage{amsmath}
\usepackage{float}
\usepackage{graphicx}
\usepackage[misc,geometry]{ifsym}
\usepackage{tabularx}
\usepackage{xtab}
\usepackage{longtable}
\usepackage{graphicx}

\usepackage{textcomp}
\usepackage{array,ragged2e}
\usepackage{pgfplots}
\usepackage{url}
\usepackage{lineno,hyperref}
\hypersetup{
  linkcolor  = red!60!black,
  citecolor  = blue!90!white,
  urlcolor   = violet!60!black,
  colorlinks = true,
}
\usepackage{orcidlink}

\usepackage{hyperref}
\usepackage{svg}
\usepackage{xcolor}
\usepackage[dvipsnames]{xcolor}
\newcommand{\orcid}[1]{\href{https://orcid.org/#1}{\includesvg[width=10pt]{orcid.svg}}}

\usepackage{threeparttable}

\usepackage{pgf-pie}  

\usepackage{longtable}
\usepackage{enumitem}

\begin{document}
\title{An Efficient Hybrid Deep Learning Approach for Detecting Online Abusive Language}
\titlerunning{Detecting Online Abusive Language}

\author{
Vuong M. Ngo\orcidlink{0000-0002-8793-0504}\inst{1}
\and
Cach N. Dang\orcidlink{0000-0001-6979-9197}\inst{2}~\textsuperscript{\Letter}
\and
Kien V. Nguyen\orcidlink{0009-0003-9804-4094}\inst{3}
\and\\
Mark Roantree\orcidlink{0000-0002-1329-2570}\inst{4}
}
\authorrunning{Ngo, V.M. et al.}
\institute{
Faculty of Information Technology, Ho Chi Minh City Open University, \\ Ho Chi Minh City, Vietnam \and
BRIDGE Research Group, Ho Chi Minh City University of Transport, \\ Ho Chi Minh City, Vietnam \and
Ho Chi Minh University of Banking, Ho Chi Minh City, Vietnam \and
Insight Centre for Data Analytics, School of Computing, Dublin City University, Dublin, Ireland\\
\email{vuong.nm@ou.edu.vn, cach@ut.edu.vn, kiennv\_htttql@hub.edu.vn, mark.roantree@dcu.ie}
}

\maketitle

\begin{abstract}
The digital age has expanded social media and online forums, allowing free expression for nearly 45\% of the global population. Yet, it has also fueled online harassment, bullying, and harmful behaviors like hate speech and toxic comments across social networks, messaging apps, and gaming communities. Studies show 65\% of parents notice hostile online behavior, and one-third of adolescents in mobile games experience bullying. A substantial volume of abusive content is generated and shared daily, not only on the surface web but also within dark web forums. Creators of abusive comments often employ specific words or coded phrases to evade detection and conceal their intentions. 
To address these challenges, we propose a hybrid deep learning model that integrates BERT, CNN, and LSTM architectures with a ReLU activation function to detect abusive language across multiple online platforms, including YouTube comments, online forum discussions, and dark web posts. The model demonstrates strong performance on a diverse and imbalanced dataset containing 77,620 abusive and 272,214 non-abusive text samples (ratio 1:3.5), achieving approximately 99\% across evaluation metrics such as Precision, Recall, Accuracy, F1-score, and AUC. This approach effectively captures semantic, contextual, and sequential patterns in text, enabling robust detection of abusive content even in highly skewed datasets, as encountered in real-world scenarios.
\end{abstract}


\section{Introduction}
\label{sec:Introduction}
The rise of the digital age and the Internet has fueled widespread use of platforms like social media and forums, enabling free expression. Nearly 45\% of the global population uses social media, often becoming addicted. However, this has also led to harassment, bullying, and harmful behavior, such as hate speech, toxic comments, and sharing obscene content. Digital technologies make such behavior possible across various online platforms, including social media, messaging, and gaming sites \cite{Chinivar:2023}. Social media has become a major platform for hostile behavior, with 65\% of parents worldwide acknowledging its prevalence. Additionally, one-third of adolescents who play mobile games have reported being victims of bullying \cite{zuckerman2020cyberbullying}. According to \cite{Petrosyan2025}, 44\% of internet users in the United States have personally encountered various forms of online harassment, with 28\% experiencing severe hostility.

Automatically detecting and analyzing online abuse text presents significant challenges due to the complexity of language, contextual ambiguity, the dynamic evolution of terminology, and the sheer volume of data. These challenges are further amplified in the detection of Abuse Material shared on the dark web, where privacy and anonymity are prioritized, making it difficult to trace perpetrators \cite{ngo2023csam}. Additionally, offenders often employ sophisticated evasion techniques, such as the use of code words, slang, cryptic abbreviations, and other forms of linguistic obfuscation, to bypass detection mechanisms and conceal illicit activities \cite{Ngo.APWG.2024}. Furthermore, legal and ethical constraints necessitate careful handling of sensitive data, limiting the extent to which automated systems can operate effectively. These factors collectively make the development of robust AI-driven detection methods both critical and highly challenging \cite{Ngo.RIVF.2022}.

Machine learning (ML) and deep learning (DL) techniques have been widely adopted to develop abusive text detection models, including Support Vector Machines (SVM) \cite{Muneer2020}, Random Forest \cite{Amrit.2017}, Decision Trees \cite{Talpur.2020}, Naïve Bayes (NB) \cite{mckeever2023determining}, K-Nearest Neighbors (KNN) \cite{Wadud.2022}, Long Short-Term Memory (LSTM) \cite{Chadaga-Springer:2023}, Bi-directional LSTM (Bi-LSTM) \cite{Li.2025}, Convolutional Neural Networks (CNN) \cite{marshan2023comparing}, and Bidirectional Encoder Representations from Transformers (BERT) \cite{NGO-CAN-2024}. However, each individual ML/DL technique has limitations, such as sensitivity to noise, difficulty handling large-scale data, or an inability to capture complex relationships. Hybrid models that combine ML/DL techniques can leverage the strengths of multiple algorithms to address these weaknesses \cite{dang2021hybrid}. Therefore, we propose a hybrid DL model combining BERT, CNN, LSTM, and ReLU activation. This integration improves prediction accuracy, especially for complex or imbalanced datasets, enhances generalization, reduces overfitting, and adapts to a wider range of tasks. These benefits are particularly valuable in domains like abusive text detection, where language can be highly variable and nuanced.

\vspace{0.8mm}
The contributions of our research can be articulated as follows:
\vspace{-1mm}
\begin{itemize}

\item Building an integrated dataset of 77,620 abusive and 272,214 non-abusive text samples from three sources to create a diverse, imbalanced collection that enables comprehensive model evaluation across real-world scenarios.

\item Proposing a hybrid DL model that effectively captures abusive text in both English and Romanized scripts by integrating BERT, CNN, LSTM, and the ReLU activation function. The model achieves very high performance with a Precision of 0.991, Recall of 0.986, Accuracy of 0.995, F1-score of 0.989, and AUC of 0.992, as evaluated using five-fold cross-validation.

\item Conducting a comprehensive comparison of the proposed model with traditional ML and standalone DL baselines using a diverse benchmark dataset comprising YouTube comments, forum discussions, and dark web posts, evaluated through 5-fold cross-validation across multiple performance metrics.
\end{itemize}

The structure of this paper is as follows. In Section \ref{sec:RW}, we review relevant research. Section \ref{sec:Methods} outlines our methodology, providing our hybrid DL model and its comprising modules in detail. In Section \ref{sec:Evaluation}, we describe the dataset and evaluate both our model and selected baseline models using various metrics, followed by a detailed performance analysis. Finally, Section \ref{sec:Conclusion-Future} concludes the paper and offers insights for future research directions.

\section{Related Work}
\label{sec:RW}

Statistical analysis of large datasets and CSV files helps uncover patterns and trends, enabling unbiased insights from raw data. Studies such as \cite{Mazzarello-Springer:2022} and \cite{Owusu-Addo-Elsevier:2023} utilized statistical methods to analyze abuse-related information. In \cite{Mazzarello-Springer:2022}, data analysis began with SPSS, followed by hierarchical logistic regression to assess three key outcomes: (1) sexual revictimization, (2) psychological dating violence, and (3) physical dating violence. Similarly, \cite{Owusu-Addo-Elsevier:2023} examined the prevalence and contributing factors of sexual abuse among adolescent girls during the COVID-19 pandemic. While, natural language processing techniques can be used to extract features and content from abuse chats and conversations. In \cite{Aguerri-Springer:2023}, the authors used Latent Dirichlet Allocation to detect tweets about child abuse and applied Conjunctive Analysis of Case Configurations, finding that longer tweets from users with smaller accounts, lacking URLs or images, were more likely to disclose abuse. However, these studies did not leverage the advantages of ML/DL techniques in detecting abuse texts.

Some studies have applied ML/DL techniques to classify abusive posts, such as \cite{Gangwar-Elsevier:2021}, \cite{Kissos-Elsevier:2020},  \cite{Laranjeira-ACM:2022}, \cite{Iftikhar.Wiley.2024} and \cite{Hole-Others:2022}. However, \cite{Gangwar-Elsevier:2021}, \cite{Kissos-Elsevier:2020} and \cite{Laranjeira-ACM:2022} primarily focused on self-figure drawings or pornographic images, utilizing object categories, visual attention mechanisms, gender-related visual features (e.g., long hair, dresses), and image metrics (e.g., luminance, sharpness).  In contrast, \cite{Iftikhar.Wiley.2024} and \cite{Hole-Others:2022} worked with videos, combining facial recognition with other biometric modalities, such as speaker recognition and age estimation.

Similar to our task, several studies have applied ML/DL techniques to detect abusive text, such as \cite{Cook-ACM:2023}, \cite{Puentes-IEEE:2023}, \cite{Chadaga-Springer:2023}, \cite{Kaur.2024}, \cite{Mosa.2025} and \cite{Li.2025}. In \cite{Cook-ACM:2023}, the authors applied natural language inference alongside the BERT model to detect harmful communication strategies. They used a dataset of 6,771 chat messages sent by child sex offenders, sourced from platforms such as MySpace and Yahoo Instant Messenger. In \cite{Puentes-IEEE:2023}, the authors proposed a model based on BERT, combined with a text tokenization method, to classify complaints across multiple dimensions and provide deeper insights into the dynamics of abuse. They analyzed 1,196 reports collected from the Colombian Child Hotline, covering topics such as grooming, sexual content disclosure, and cyberbullying. Meanwhile, \cite{Chadaga-Springer:2023} employed explainable artificial intelligence (XAI) techniques to identify early warning signs, aiming to raise awareness among individuals with limited prior knowledge of child sexual abuse. The study was based on responses to survey questions collected from 3,002 participants. In \cite{Kaur.2024}, Kaur et al. developed a robust abusive language detection model using a dataset of 14,200 English tweets. The approach leveraged DL architectures, specifically LSTM and Gated Recurrent Unit (GRU) networks, to capture contextual dependencies and sequential patterns for accurate classification.

Nearly, \cite{Mosa.2025} introduced a hybrid feature selection approach combining the Enhanced Non-Dominated Sorting Genetic Algorithm II with XGBoost, aiming to optimize classification performance while reducing the number of selected features. The model was evaluated on the Levantine Hate Speech and Abusive dataset, consisting of 5,846 political Arabic tweets from Syria and Lebanon, labeled as normal, abusive, or hate speech. Finally, in \cite{Li.2025}, the authors applied feature selection techniques such as XGBoost and Multilayer Perceptron to enhance a BiLSTM model for detecting cyberbullying texts and identifying specific bullying types. The dataset comprised 15,294 microblogs containing personal cyberbullying attacks, categorized accordingly, along with 17,826 non-cyberbullying texts for comparison. 

However, the ML/DL models for abusive text detection proposed in the aforementioned studies (i.e., \cite{Cook-ACM:2023}, \cite{Puentes-IEEE:2023}, \cite{Chadaga-Springer:2023}, \cite{Kaur.2024}, \cite{Mosa.2025}, and \cite{Li.2025}) differ from ours in both scope and methodological approach.
\section{Methodology}
\vspace{-1mm}
\label{sec:Methods}

\vspace{-2mm}
\begin{figure}[H]
    \centering
    \includegraphics[width=1\linewidth, height = 4.4cm]{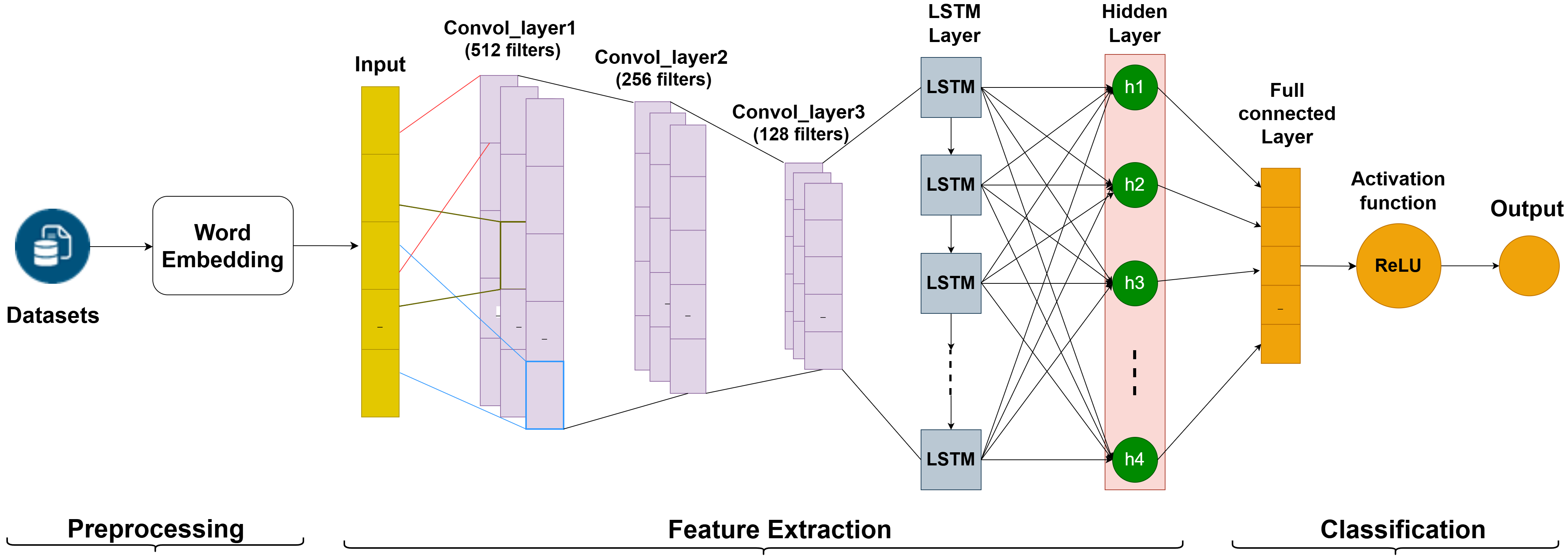}
    \caption{Process of methodology for hybrid model}
    \label{fig:our-model-architecture}
\end{figure}
\vspace{-3mm}

In this paper, we propose a hybrid DL model for detecting online abusive comments. The model architecture, illustrated in Figure \ref{fig:our-model-architecture}, is designed to predict the abuse polarity of a text and classify it accordingly. It consists of three main modules: {\tt Preprocessing}, {\tt Feature Extraction} and {\tt Classification}. 
In the {\tt Preprocessing} module, BERT is employed as the word embedding model to generate feature vectors. The {\tt Feature Extraction} module integrates CNN and LSTM models to capture both local and sequential features of the text. Finally, the {\tt Classification} module applies a ReLU activation function to produce the final output.

The purpose of this module combination is to enhance detection accuracy compared to single-model approaches when applied to complex data from diverse sources, although it requires longer computation time. The hybrid architecture leverages the complementary strengths of BERT, CNN, and LSTM: BERT captures the full contextual meaning of words through bidirectional analysis, CNN effectively extracts salient textual features, and LSTM retains past information via its cell states, enhancing the model’s ability to learn sequential dependencies.

\subsection{Word Embedding Layer}
The initial phase of our model involves transforming the input text into a sequence of word embeddings—dense vector representations that encode semantic relationships among words. In this work, we deployed a pre-trained BERT model tailored for the Online Abusive Comments (OAC) dataset. After fine-tuning its parameters, BERT serves as a feature extractor, generating contextualized embeddings for the proposed hybrid framework. The OAC dataset is passed through the BERT model to generate feature vectors, which are then fed into the subsequent layers of the hybrid architecture. The BERT-base-uncased architecture was used to produce contextualized embeddings with a hidden size of 768.

\subsection{Convolutional Neural Networks model}

\begin{figure}[H]
    \centering
    \includegraphics[width=1\linewidth]{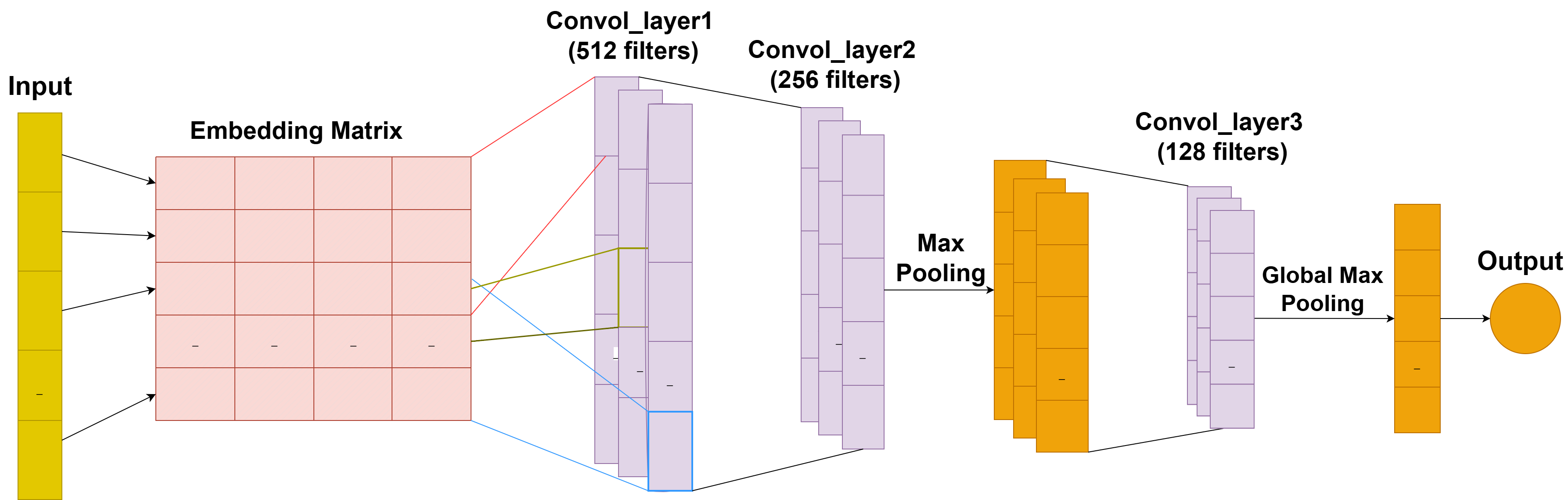}
    \caption{A convolutional neural network}
    \label{fig:CNN}
\end{figure}
\vspace{-2mm}

A Convolutional Neural Network (CNN) is a form of feedforward neural network that processes data sequentially from input to output, without any feedback loops. It employs a deep layered architecture \cite{yamashita2018convolutional}, usually starting with convolutional and pooling (or subsampling) layers that extract and transform features from the input data. These are then followed by one or more fully connected layers that perform the final classification.

While the general architecture of CNNs is independent of the dimensionality of the input data, their implementation depends on it. The dimensionality determines how subsampling filters slide across the data. In natural language processing, a one-dimensional convolutional layer (1D CNN) with m filters is commonly applied, producing an m-dimensional feature vector for each document n-gram. These feature vectors are then aggregated using max-pooling, followed by a ReLU (Rectified Linear Unit) activation function. The resulting output is then forwarded to a fully connected (linear) layer, which performs the ultimate classification step, as illustrated in Figure \ref{fig:CNN}. To capture local contextual patterns in BERT embeddings, a sequence of three 1D convolutional layers with 512, 256, and 128 filters, respectively, and a kernel size of 3 is applied.

Let \( w_{i:n} \in \mathbb{R}^d \) denote the input text consisting of \( n \) words, where each word is represented as a \( d \)-dimensional embedding vector. The sequence of embeddings forms a  \( d \times n \)  matrix that serves as the input to the convolutional layer, where filters are applied across the text.

For each \( l \)-word \( n \)-gram, we define:
\begin{equation}
   c_i = [w_i, \dots, w_{i+l-1}] \in \mathbb{R}^{(d \times l)}, \quad 0 \leq i \leq n - l
\end{equation}

For each filter \( f_j \in \mathbb{R}^{d \times l} \), 
the convolution operation is computed as the inner product \( \langle c_i, f_j \rangle \). The resulting convolutional feature maps are collected into a matrix \( F \in \mathbb{R}^{n \times m} \), where 
\( m\) is the number of filters.

A max-pooling operation is then applied across the 
\( n \)-gram dimension:
\begin{equation}
    p_j = \max\limits_{i} (F_{ij})
\end{equation}

The pooled features are passed through a ReLU non-linearity to introduce activation.

Finally, the extracted feature representation is passed through a fully connected layer that produces a probability distribution across the target classes. The class corresponding to the highest probability is then chosen as the final prediction.

\subsection{Long short-term memory model}

\begin{figure}[H]
    \centering
    \includegraphics[width=1\linewidth, height = 6cm]{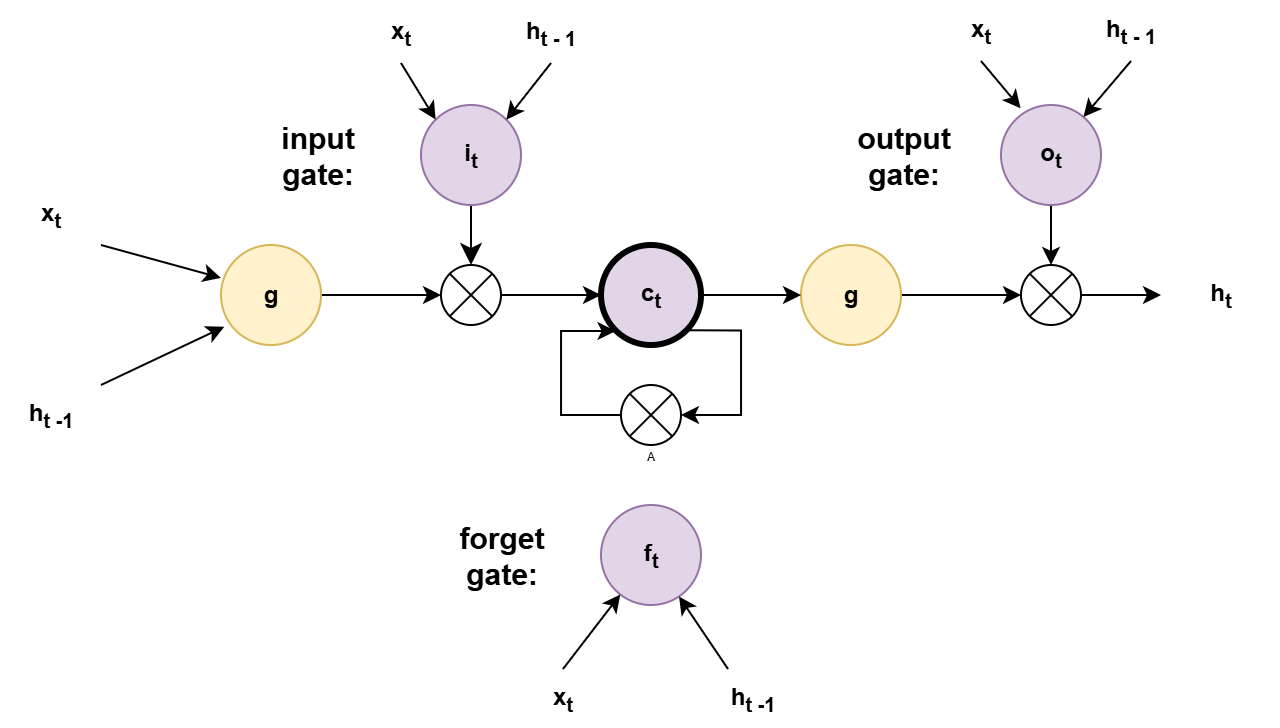}
    \caption{Illustration of a LSTM block \cite{palangi2016deep}}
    \label{fig:LSTM-block}
\end{figure}

The Long Short-Term Memory (LSTM) network is a specialized variant of the Recurrent Neural Network architecture \cite{palangi2016deep}. Each LSTM unit comprises three main gates—the {\tt forget gate}, {\tt input gate}, and {\tt output gate}—as well as input/output components and a memory cell that allows the model to learn and retain long-term dependencies. The structure of an LSTM block is illustrated in Figure \ref{fig:LSTM-block}. In our model, the extracted features are passed through an LSTM layer with a hidden size of 500 to effectively capture sequential dependencies.

The {\tt forget gate} regulates the extent to which information from the previous cell state $c_{t-1}$ is preserved:
\begin{equation}
f_t = \sigma \left( W_f \cdot \left[ h_{t-1}, x_t \right] + b_f \right)
\end{equation}

The {\tt input gate} determines the amount of new information to incorporate into the cell state, while the candidate cell state 
\( \tilde{c}_t \) is calculated as follows:
\begin{equation}
i_t = \sigma(W_i [h_{t-1}, x_t] + b_i), \quad 
\tilde{c}_t = \tanh(W_c [h_{t-1}, x_t] + b_c)
\end{equation}

\vspace{-3mm}
\vspace{-2mm}

The updated cell state is then:
\begin{equation}
c_t = f_t \cdot c_{t-1} + i_t \cdot \tilde{c}_t
\end{equation}

Finally, the {\tt output gate} controls the proportion of the cell state that is exposed to the hidden output:
\begin{equation}
    o_t = \sigma(W_o [h_{t-1}, x_t] + b_o), \quad 
    h_t = o_t \cdot \tanh(c_t)
\end{equation}

Each gate has its own learnable weights and biases, enabling the network to determine how much past information to retain, how much new input to incorporate, and how much of the internal state to expose at each time step.

\vspace{-1mm}
\subsection{Fully Connected Layer and Output}
The context vector produced by the LSTM layer with an attention mechanism is fed into a fully connected (dense) layer utilizing a ReLU activation function. This layer enhances the high-level features learned by the CNN and LSTM modules, enabling the model to capture more intricate and non-linear relationships within the data.

Next, the output from the dense layer is passed to the final output layer, which employs an appropriate activation function (e.g., softmax) to generate the abuse classification probabilities. The model is optimized using the categorical cross-entropy loss function, and the class label with the highest probability is selected as the final prediction.

\section{Model Evaluation}
\label{sec:Evaluation}

\subsection{Our Dataset}
\begin{figure}[H]
    \centering
    \captionsetup{justification=centering}
	\begin{center}
        \begin{tabular}{cc}
          \includegraphics[width=0.48\textwidth, height=2.9cm]{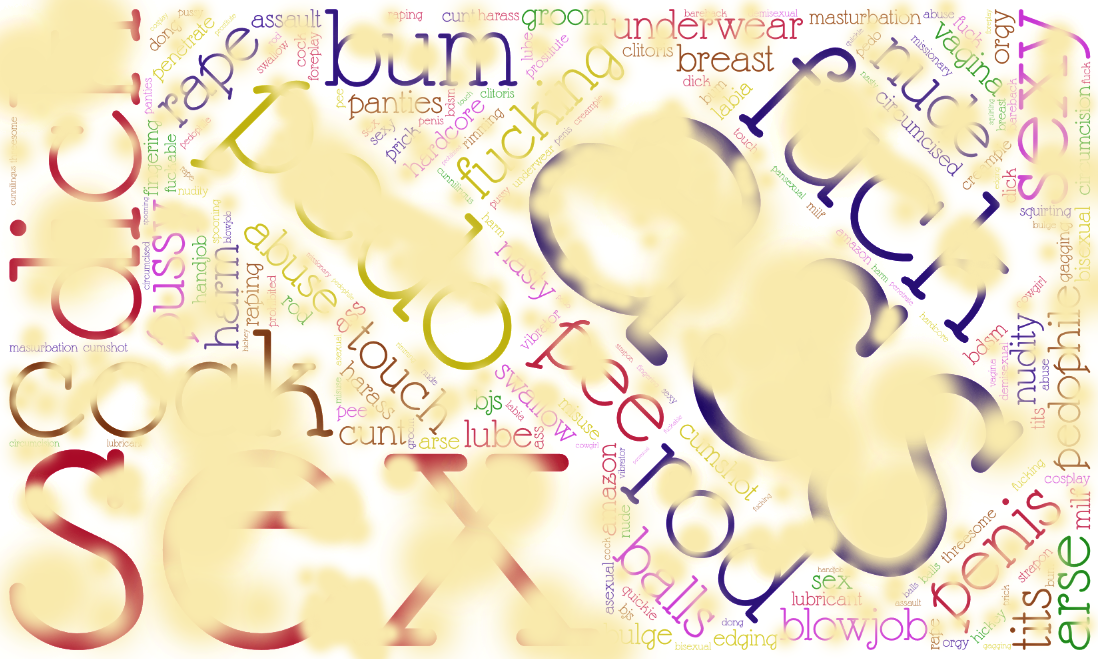} & \hspace{2mm}
          \includegraphics[width=0.48\textwidth, height=2.9cm]{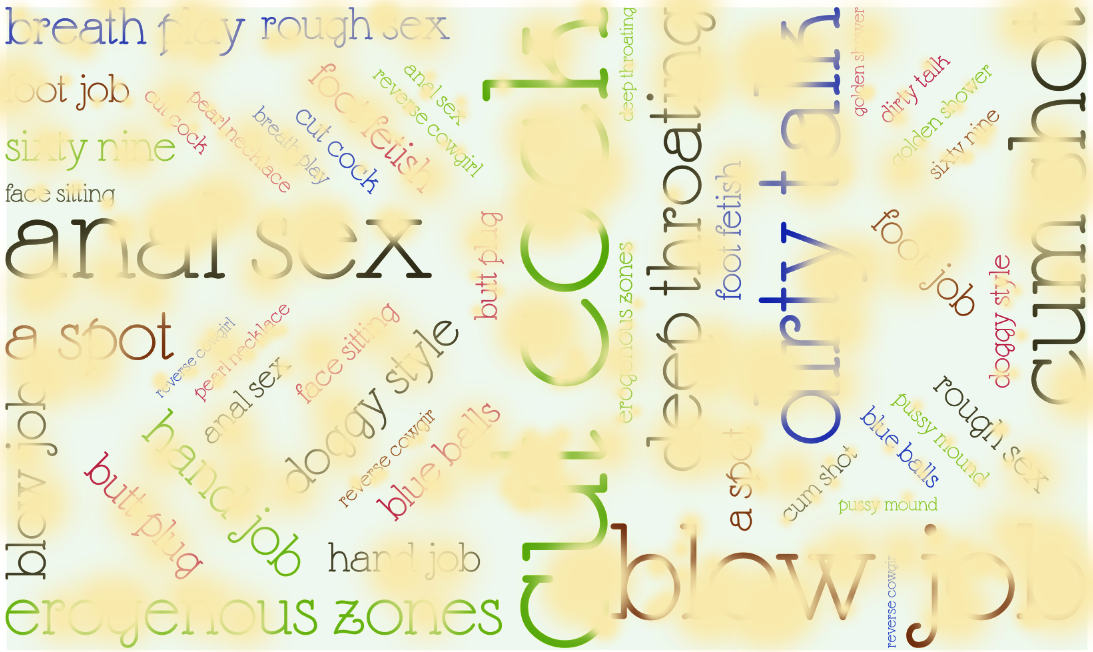}
        \end{tabular}       
	\end{center}
    \vspace{-4mm}
  	\caption{Phrases related to sexual abuse on the dark web}
    \vspace{-3mm}
	\label{fig_sap}
\end{figure}

To evaluate model performance, our dataset includes 77,620 abusive and 272,214 non-abusive text samples, integrated from three datasets. This diverse and imbalanced collection reflects real-world conditions, where abusive content is considerably less frequent than normal content. It enhances the model’s generalization across different contexts, supports robust evaluation under realistic class distributions, and enables fair comparison with other imbalanced benchmark datasets. The combined datasets include:
\vspace{-2mm}
\begin{enumerate}
\item Darkweb Dataset: We created the Darkweb dataset comprising 4,600 samples extracted from 352,000 dark web forum posts collected in 2022. It includes 2,500 child sexual abuse (CSA)-related and 2,100 non-CSA samples. Among them, 2,000 CSA and 100 non-CSA samples contain at least one sexual abuse phrase, see more details in \cite{Ngo.APWG.2024}. Figure \ref{fig_sap} presents blurred word clouds depicting single words and two-word phrases associated with sexual abuse, which were extracted from the post contents.

\item PAN12 Dataset\footnote{\url{https://pan.webis.de/clef12/pan12-web/}}: Developed for the CLEF 2012 competition, this dataset aims to detect predatory behavior in online chats. It includes 198,054 conversations, consisting of 4,029 abusive and 194,025 non-abusive samples.

\item Roman Urdu Dataset: Contains 147,180 YouTube comments, evenly divided into abusive (73,590) and non-abusive (73,590) classes \cite{Akhter-IEEE:2020}.
\end{enumerate}

\vspace{-4mm}

\subsection{Experimental Design and Metrics}

To assess the model's performance, we use accuracy (ACC), precision (P), recall (R), and F1-score (F1), which are computed from the confusion matrix values: true positives (TP), false positives (FP), true negatives (TN), and false negatives (FN). These metrics are defined as follows: accuracy is computed by $ACC = \frac{TP + TN}{TP + FP + TN + FN}$, precision by $P = \frac{TP}{TP + FP}$, recall by $R = \frac{TP}{TP + FN}$, and the F1-score by $F1 = \frac{2 \cdot P \cdot R}{P + R}$.

The AUC (Area Under the Curve) metric is also used to assess a model’s overall classification performance. It represents the area beneath the ROC curve, which graphs the True Positive Rate against the False Positive Rate at various threshold settings. Higher AUC values, approaching 1, signify greater classification accuracy and model effectiveness. 



The experiments were carried out in a Kaggle Notebook environment featuring an NVIDIA Tesla T4 GPU with 16 GB of memory, of which up to 15 GB was available per session, alongside an Intel(R) Xeon(R) CPU @ 2.00 GHz with four logical cores. The setup used Python 3.10, with the PyTorch 2.5 framework and Hugging Face’s Transformers library for implementing and fine-tuning the BERT-based model. The environment also provided up to 30 GB of RAM and 57.6 GB of temporary storage, which proved sufficient for training, validating, and evaluating the model’s performance efficiently. We use 5-fold cross-validation during model development where the reported classification performance and execution times correspond to the averaged scores across the five runs.

\subsection{Results and Discussion}

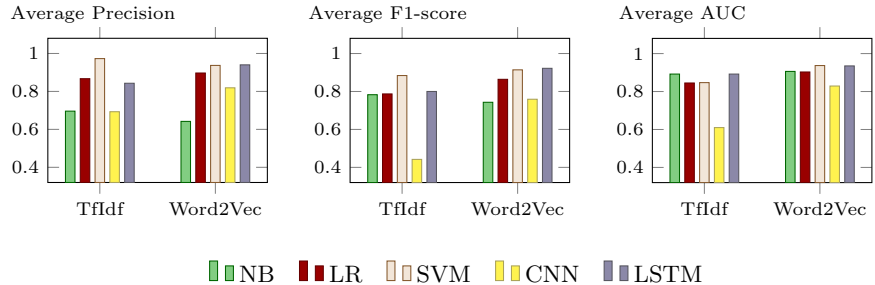
\begin{figure}[H]
\small
 \begin{center}
  \begin{tikzpicture}
   \pgfplotsset{height=35mm, width= 45mm, xlabel near ticks, ylabel near ticks, 
    plot 1/.style={green!40!black,fill={rgb,255:red,130; green,205; blue,130},mark=none},%
    plot 2/.style={red!40!black,fill=red!60!black,mark=none},%
    plot 3/.style={brown!70!black,fill=brown!20!white,mark=none},%
    plot 4/.style={yellow!60!black,fill=yellow!80!white,mark=none},%
    plot 7/.style={CadetBlue!30!black,fill=CadetBlue!80!white,mark=none},%
   }

   \begin{axis}[
    name=left axis,
    font=\scriptsize,
    ybar,
    bar width=3.6pt,
    enlargelimits=0.45,
    legend style={/tikz/every even column/.append style={column sep=0.25cm},
    draw=none, fill=none, font=\footnotesize, legend columns=-1, anchor = south, xshift= 26mm, yshift= -34mm}, 
    y label style={align=center},
    ylabel={Average Precision},
    every axis y label/.style={
    	at={(ticklabel* cs:1.05)},
    	anchor=south, xshift= 6mm,
    },
    ymin=0.5, ymax=0.9,
    symbolic x coords={TfIdf, Word2Vec},
    xtick=data,
    every node near coord/.append style={font=\scriptsize},  
    ]
    \addplot[plot 1] coordinates{(TfIdf,0.696) (Word2Vec,0.642)};
    \addplot[plot 2] coordinates{(TfIdf,0.867) (Word2Vec,0.897)};
    \addplot[plot 3] coordinates{(TfIdf,0.973) (Word2Vec,0.937)};
    \addplot[plot 4] coordinates{(TfIdf,0.693) (Word2Vec,0.819)};
    \addplot[plot 7] coordinates{(TfIdf,0.843) (Word2Vec,0.940)};

	\legend{NB, LR, SVM, CNN, LSTM}
   \end{axis}
   
   \begin{axis}[
    name=center axis,
    at=(left axis.east),
    xshift= 1.1cm,
	yshift= -0.96cm,
    font=\scriptsize,
    ybar,
    bar width=3.6pt,
    enlargelimits=0.45,
    y label style={align=center},
    ylabel={Average F1-score},
    every axis y label/.style={
    	at={(ticklabel* cs:1.05)},
    	anchor=south, xshift= 5mm,
    },
    ymin=0.5, ymax=0.9,
    symbolic x coords={TfIdf, Word2Vec},
    xtick=data,
    every node near coord/.append style={font=\scriptsize},  
    ]
    \addplot[plot 1] coordinates{(TfIdf,0.783) (Word2Vec,0.743)};
    \addplot[plot 2] coordinates{(TfIdf,0.787) (Word2Vec,0.864)};
    \addplot[plot 3] coordinates{(TfIdf,0.884) (Word2Vec,0.914)};
    \addplot[plot 4] coordinates{(TfIdf,0.442) (Word2Vec,0.759)};
    \addplot[plot 7] coordinates{(TfIdf,0.80) (Word2Vec,0.922)};

   \end{axis}
   
   \begin{axis}[
    name=right axis,
    at=(center axis.east),
    xshift= 1.1cm,
	yshift= -0.96cm,
    font=\scriptsize,
    ybar,
    bar width=3.6pt,
    enlargelimits=0.45,
    y label style={align=center},
    ylabel={Average AUC},
    every axis y label/.style={
    	at={(ticklabel* cs:1.05)},
    	anchor=south, xshift= 4mm,
    },
    ymin=0.5, ymax=0.9,
    symbolic x coords={TfIdf, Word2Vec},
    xtick=data,
    every node near coord/.append style={font=\scriptsize},	
    ]
    \addplot[plot 1] coordinates{(TfIdf,0.892) (Word2Vec,0.906)};
    \addplot[plot 2] coordinates{(TfIdf,0.845) (Word2Vec,0.903)};
    \addplot[plot 3] coordinates{(TfIdf,0.847) (Word2Vec,0.937)};
    \addplot[plot 4] coordinates{(TfIdf,0.610) (Word2Vec,0.829)};
    \addplot[plot 7] coordinates{(TfIdf,0.892) (Word2Vec,0.935)};
   \end{axis}
   
  \end{tikzpicture}

 \end{center}
 \captionsetup{justification=centering}
 \vspace{-4mm}
 \caption{Average Precision, F1-score and AUC of baseline ML and DL models}
 \label{fig:ml_dl_models}
 \vspace{-4mm}
\end{figure}

The average Precision, F1-score, and AUC of the NB, LR, SVM, CNN, and LSTM models using TF-IDF ($tf.idf$) and Word2Vec ($w2v$) feature representations are presented in Figure~\ref{fig:ml_dl_models}, averaged across five-fold cross-validation. Overall, models leveraging Word2Vec embeddings generally outperform their TF-IDF counterparts across most evaluation metrics, reflecting the benefits of semantic representations in capturing contextual meaning. The only exception is the NB model, where NB$_{tf.idf}$ achieves higher Precision (0.696) and F1-score (0.783) than NB$_{w2v}$, although the latter records a slightly higher AUC (0.906 vs. 0.892), indicating a marginally better discriminative ability.

\vspace{-2mm}

Among the remaining models, LR$_{w2v}$, SVM$_{w2v}$, CNN$_{w2v}$, and LSTM$_{w2v}$ consistently achieve superior results. Specifically, LR$_{w2v}$ outperforms LR$_{tfidf}$ in Precision (0.897 vs. 0.867), F1-score (0.864 vs. 0.787), and Accuracy (0.903 vs. 0.845). SVM$_{w2v}$ also shows improvements over SVM${tfidf}$ in F1-score (0.914 vs. 0.884) and AUC (0.937 vs. 0.847). CNN$_{w2v}$ also surpasses CNN$_{tfidf}$ across all three evaluation metrics (Precision: 0.819 vs. 0.693, F1-score: 0.759 vs. 0.442, AUC: 0.829 vs. 0.610), demonstrating the advantage of semantic embeddings. Similarly, LSTM$_{w2v}$ achieves the best overall performance among all models, with a Precision of 0.94, F1-score of 0.922, and AUC of 0.935. These results highlight the effectiveness of DL and kernel-based models when combined with Word2Vec embeddings. Therefore, the best-performing models in this comparison are NB$_{tf.idf}$, LR$_{w2v}$, SVM$_{w2v}$, CNN$_{w2v}$, and LSTM$_{w2v}$.

\begin{figure}[H]
\begin{center}
\scriptsize
\begin{tabular}{ccccc}
    \includegraphics[width=.27\textwidth]{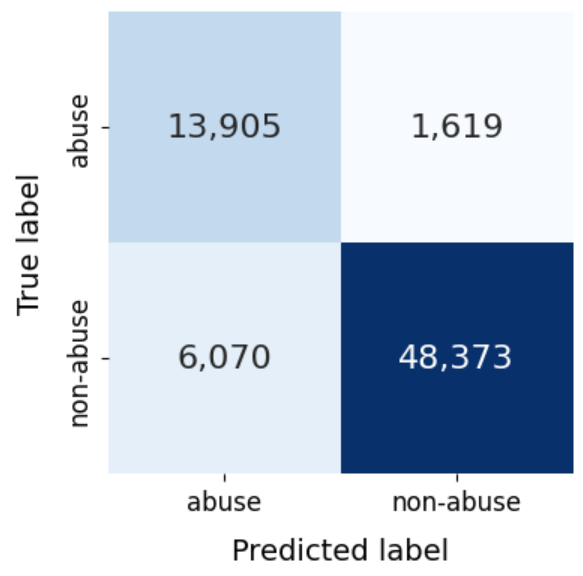} \hspace{6mm}&&
    \includegraphics[width=.27\textwidth]{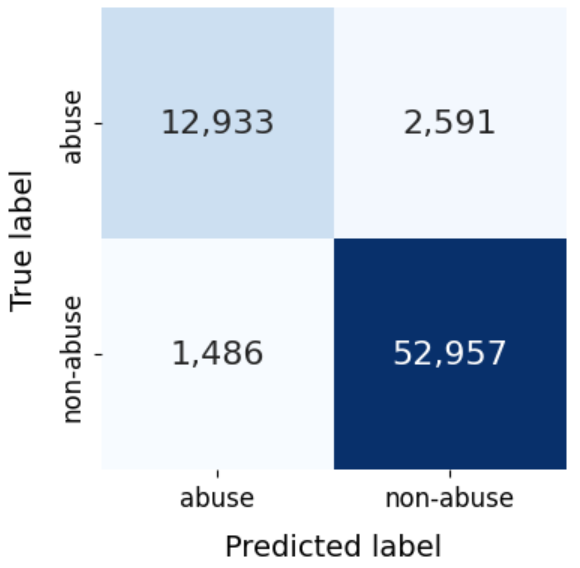} \hspace{6mm}&&
    \includegraphics[width=.368\textwidth]{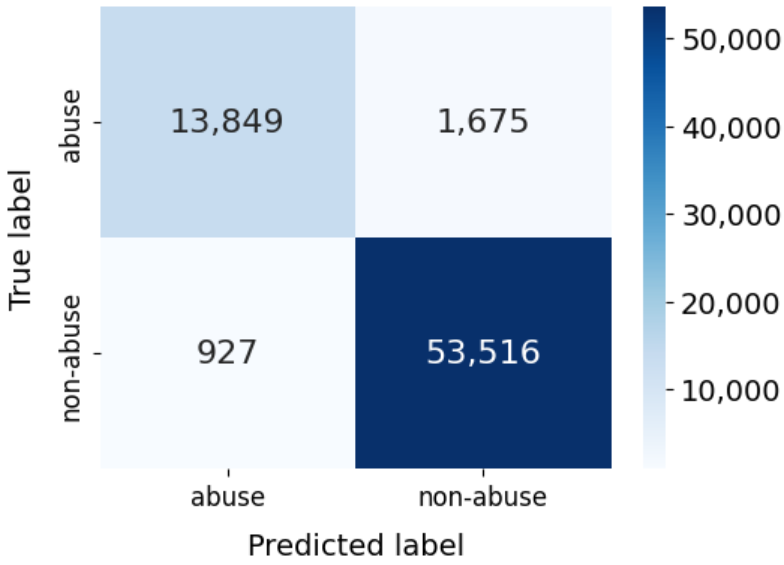}
    \vspace{2mm}
    \\
    \hspace{4mm} (a) NB$_{tf.idf}$  && \hspace{3 mm} (b) LR$_{w2v}$  && \hspace{-5mm} (c) SVM$_{w2v}$\\
\end{tabular}

\vspace{5mm}

\begin{tabular}{ccccccc}
    \includegraphics[width=.234\textwidth]{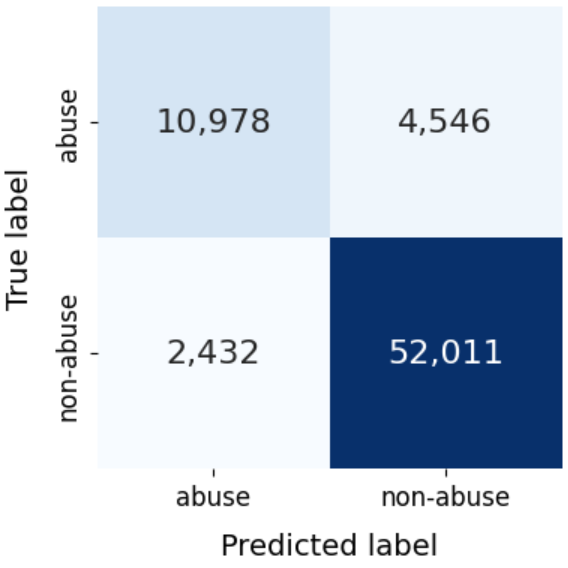} &&
    \includegraphics[width=.234\textwidth]{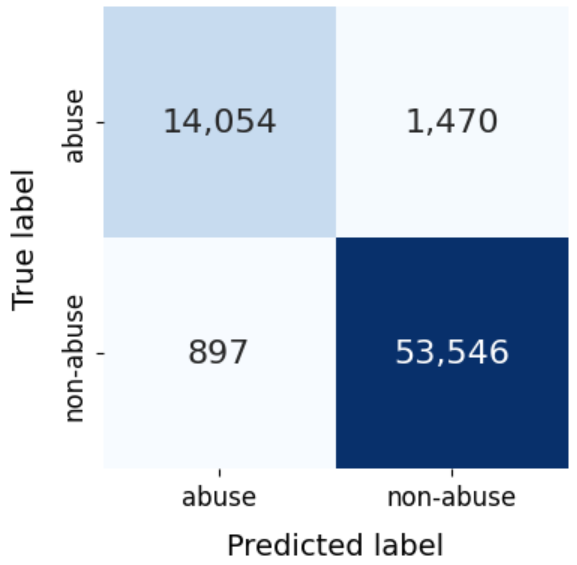} &&
    \includegraphics[width=.234\textwidth]{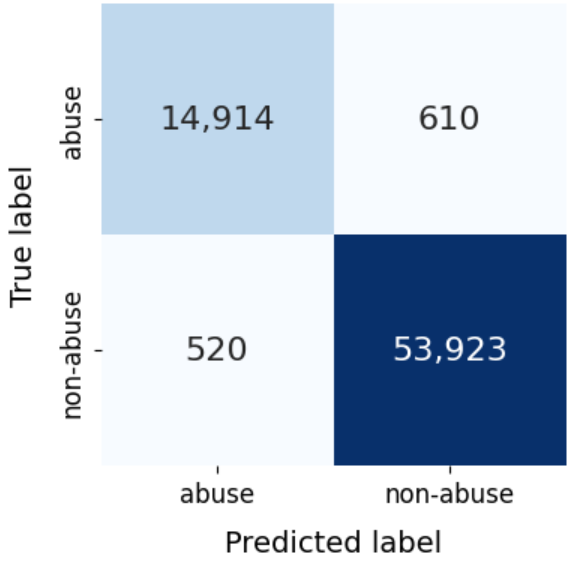} &&
    \includegraphics[width=.234\textwidth]{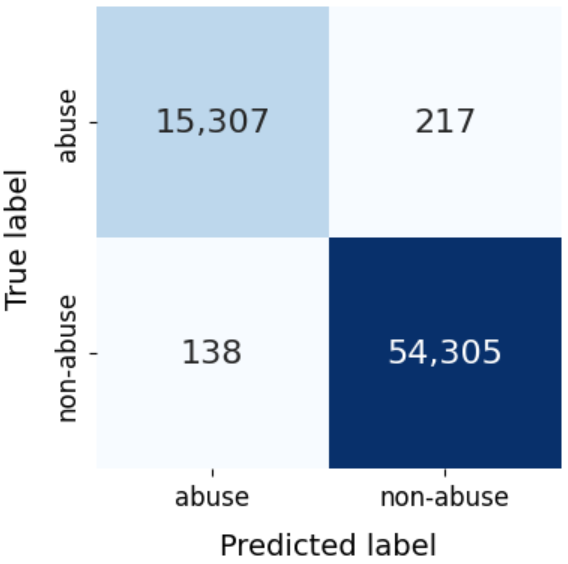}
    \vspace{2mm}
    \\
    \hspace{5mm}(d) CNN$_{w2v}$ && \hspace{6mm}(e) LSTM$_{w2v}$ && \hspace{4mm}(f) BERT && \hspace{6mm}(g) Our Model\\
\end{tabular}
\caption{Confusion Matrices for models}
\label{fig:confusion-matrices}
\end{center}
\end{figure}

Figure \ref{fig:confusion-matrices} presents and conpares the confusion matrix components  across  our proposed hybrid model, BERT and the the best-performing ML/DL models based on the $tf.idf$ or $w2v$. The proposed hybrid model achieved the highest number of TP (15,307) and TN (54,305), indicating its superior ability to correctly identify both abusive and non-abusive comments. It also recorded the lowest FP (138) and FN (217), demonstrating a strong balance between precision and recall. In contrast, traditional models such as NB and LR showed weaker performance, with significantly higher FP and FN. For example, NB$_{tf.idf}$ produced 6,070 FP and 1,619 FN, while CNN$_{w2v}$ also struggled with 4,546 FN. BERT and LSTM$_{w2v}$ performed considerably better, reducing misclassifications and achieving stronger overall accuracy; however, our model consistently outperformed them in all four metrics.

\vspace{-2mm}
\vspace{-6mm}

Table \ref{tab:classification-performance} summarizes the experimental results across all models. The proposed hybrid model achieved the best overall performance, with the highest Precision (0.991), Recall (0.986), Accuracy (0.995), F1-score (0.989), and AUC (0.992). This indicates its strong ability to correctly identify both abusive and non-abusive comments with minimal errors. BERT and LSTM$_{w2v}$ also performed competitively, achieving high precision and recall but slightly lower accuracy and F1-score compared to the hybrid model. Traditional ML models, including NB$_{tf.idf}$, LR$_{w2v}$, and SVM$_{w2v}$, delivered moderate performance, with NB showing the lowest precision (0.696) and CNN$_{w2v}$ yielding the weakest overall results. In terms of computational efficiency, NB$_{tf.idf}$ and LR$_{w2v}$ had the fastest training and prediction times, while DL models, particularly BERT and the hybrid model, required substantially longer processing times due to their complexity. Overall, the hybrid model offers the best trade-off between classification performance and reliability, albeit at the cost of increased computational time. These results highlight the effectiveness of integrating DL architectures in a hybrid framework to enhance both sensitivity and specificity in online abusive comment detection.

\vspace{-6mm}
\begin{table}[H]
 \scriptsize
 \begin{center}
  \caption{Average model performance across 5-fold cross-validation}
  \begin{threeparttable}[t]
  \begin{tabular}{|l|r|r|r|r|r|r|r|}
    \hline
        \textbf{Classification} & \multicolumn{3}{c|}{\textbf{Machine Learning}} & \multicolumn{3}{c|}{\textbf{Deep Learning}} & \textbf{Our Hybrid} \\ \cline{2-7} 
         \textbf{Performance} & \textbf{NB$_{tf.idf}$} & \textbf{LR$_{w2v}$} & \textbf{SVM$_{w2v}$} & \textbf{CNN$_{w2v}$} & \textbf{LSTM$_{w2v}$} & \textbf{BERT} 
            & \textbf{Model} \\ \hline        
        Average Precision & 0.696 & 0.897 & 0.937 & 0.819 & 0.940 & 0.966 & \textbf{0.991} \\ \hline
        Average Recall & 0.896 & 0.833 & 0.892 & 0.707 & 0.905 & 0.961 & \textbf{0.986} \\ \hline
        Average Accuracy & 0.890 & 0.942 & 0.963 & 0.900 & 0.966 & 0.984 & \textbf{0.995} \\ \hline
        \rowcolor{cyan!30} 
        Average F1-score & 0.783 & 0.864 & 0.914 & 0.759 & 0.922 & 0.964 & \textbf{0.989} \\ \hline
        \rowcolor{lime}
        Average AUC & 0.892 & 0.903 & 0.937 & 0.829 & 0.935 & 0.976 & \textbf{0.992} \\ \hline

         Avg Training time (sec) & 3.7 & 26.7 & 340.9 & 192.1 & 233.9 & 4,672.8 & 5,076.8 \\ \hline
         \rowcolor{orange!80}
         Avg Pred. time (sec) & 1.75 & 0.03 & 42.65 & 1.80 & 1.38 & 203.60 & 459.78 \\ \hline
    \end{tabular}
  \end{threeparttable}
  \label{tab:classification-performance}
 \end{center}
\end{table}
\vspace{-7mm}

\begin{figure}[H]
\begin{center}
\small
\begin{tabular}{ccc}
    \includegraphics[width=.48\textwidth]{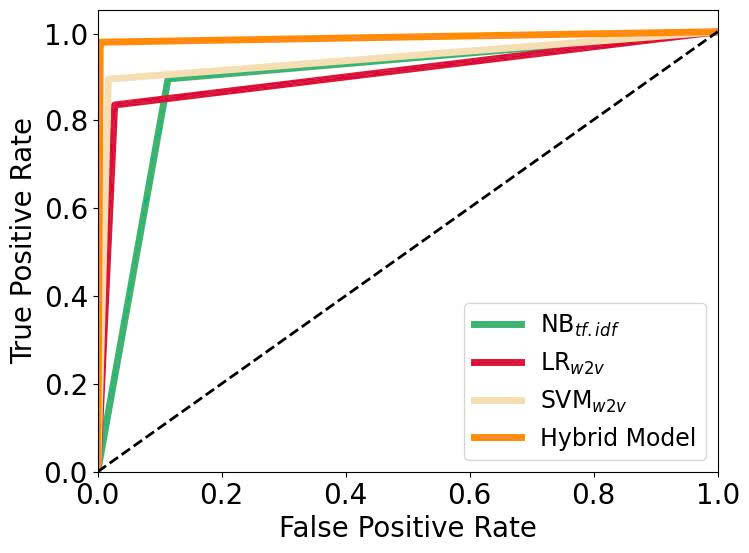} &&
    \includegraphics[width=.48\textwidth]{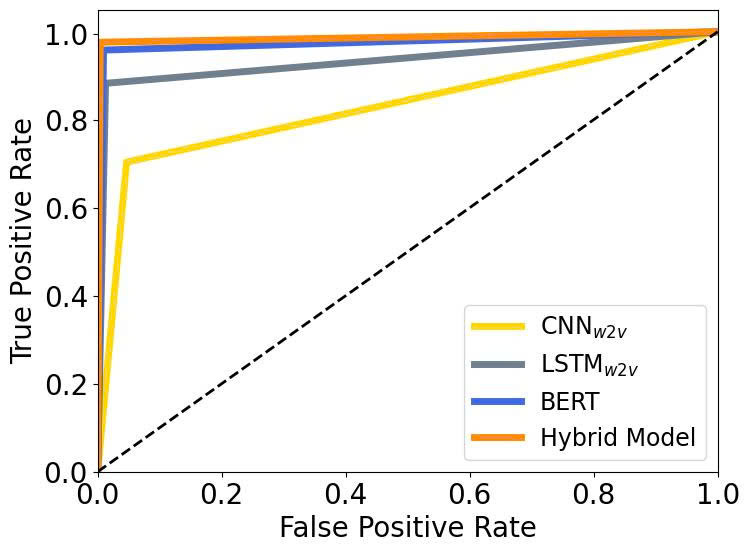} 
    \vspace{2mm}
    \\
    (a) ML models and our model && \hspace{0mm} (b) DL models and our model \\
\end{tabular}
\vspace{-1mm}
\caption{ROC curves for models}
\label{fig:roc_curves}
\end{center}
\end{figure}

Additionally, the Hybrid model also achieves the highest performance on the ROC curve, as shown in Figure \ref{fig:roc_curves}, with an averaged AUC of 99.2\%. The curve almost touches the top-left corner, indicating an optimal balance between the true positive rate and the false positive rate.

\vspace{-2mm}
\vspace{-12mm}

\newpage
\section{Conclusion and Future Work}
\label{sec:Conclusion-Future}
\vspace{-1mm}

This paper presents an efficient hybrid deep learning model that combines BERT, CNN, and LSTM architectures with a ReLU activation function to detect abusive language across multiple online platforms. By effectively integrating these components, the model captures semantic, contextual, and sequential dependencies in text, enabling robust classification of abusive and non-abusive content. Evaluation on a large, diverse, and imbalanced dataset—including YouTube comments, online forum posts, and dark web content—demonstrated its effectiveness under realistic conditions.

Experimental results showed that the proposed hybrid model outperformed traditional ML and standalone DL methods. It achieved strong results with a Precision of 0.991, Recall of 0.986, Accuracy of 0.995, F1-score of 0.989, and AUC of 0.992, confirming its high discriminative ability and robustness. Although computationally more demanding than simpler models, this trade-off remains acceptable for real-world applications requiring reliable performance. Overall, the hybrid model provides an effective and flexible approach to abuse detection, serving as a strong foundation for future research on content moderation and digital safety.

Future work will focus on extending this framework to multilingual and code-mixed text, improving interpretability through XAI methods, and incorporating multimodal features such as images or user metadata. Additionally, optimizing computational efficiency via model compression or pruning will support real-time deployment in large-scale social media monitoring systems.

\section*{Acknowledgement}
\vspace{-1mm}
This publication has emanated from research conducted with the financial support of Taighde Éireann – Research Ireland under Grant number 12/RC/2289\_P2.

\bibliographystyle{splncs04}
\bibliography{References.bib}

@inproceedings{mckeever2023determining,
  author={Mckeever, Susan and Thorpe, Christina and Ngo, Vuong M},
  title={Determining Child Sexual Abuse Posts based on Artificial Intelligence},
  booktitle={2023 International Society for the Prevention of Child Abuse \& Neglect Congress (ISPCAN-2023)},
  year={2023},
  address={Edinburgh, UK},
  pages={24--27},
}

@article{marshan2023comparing,
  author = {Marshan, Alaa and Nizar, Farah Nasreen Mohamed and Ioannou, Athina and Spanaki, Konstantina},
  title = {Comparing Machine Learning and Deep Learning Techniques for Text Analytics: Detecting the Severity of Hate Comments Online},
  journal = {Information Systems Frontiers},
  year = {2023},
}

@article{Talpur.2020,
    author = {Talpur, Bandeh Ali AND O’Sullivan, Declan},
    journal = {PLOS ONE},
    publisher = {Public Library of Science},
    title = {Cyberbullying severity detection: A machine learning approach},
    year = {2020},
    volume = {15},
    pages = {1-19},
    number = {10},
}

@article{Amrit.2017,
    title = {Identifying child abuse through text mining and machine learning},
    journal = {Expert Systems with Applications},
    volume = {88},
    pages = {402-418},
    year = {2017},
    issn = {0957-4174},
    author = {Chintan Amrit and Tim Paauw and Robin Aly and Miha Lavric},
}

@article{Wadud.2022,
    title = {How can we manage Offensive Text in Social Media - A Text Classification Approach using LSTM-BOOST},
    journal = {International Journal of Information Management Data Insights},
    volume = {2},
    number = {2},
    pages = {100095},
    year = {2022},
    issn = {2667-0968},
    author = {Md. Anwar Hussen Wadud and others},
}

@article{Muneer2020,
  author    = {Muneer, A. and Fati, S. M.},
  title     = {A comparative analysis of machine learning techniques for cyberbullying detection on twitter},
  journal   = {Future Internet},
  volume    = {12},
  number    = {11},
  year      = {2020},
}

@article{Chinivar:2023,
    title = {Online offensive behaviour in socialmedia: Detection approaches, comprehensive review and future directions},
    journal = {Entertainment Computing},
    volume = {45},
    pages = {100544},
    year = {2023},
    issn = {1875-9521},
    author = {Sneha Chinivar and others},
}

@misc{Petrosyan2025,
  author = {Ani Petrosyan},
  title = {Impact of Online Hate and Harassment in the U.S. 2020},
  year = {2025},
  publisher = {Statista},
  url = {https://www.statista.com/statistics/971876/societal-impact-of-online-hate-harassment-usa/},
  note = {Accessed: 2025-09-02}
}

@misc{zuckerman2020cyberbullying,
  author = {Arthur Zuckerman},
  title = {60 Cyberbullying Statistics: 2020/2021 Data, Insights \& Predictions},
  url = {https://comparecamp.com/cyberbullying-statistics},
  note = {Accessed: 2025-04-02}
}

@inproceedings{Ngo.APWG.2024,
 author               = {V. M. Ngo and S. Mckeever and C. Thorpe},
 year                 = {2024},
 title                = {Identifying Online Child Sexual Texts in Dark Web through Machine Learning and Deep Learning Algorithms},
 booktitle            = {the APWG.EU Technical Summit and Researchers Sync-Up (APWG.EU-Tech 2023)},
 pages                = {1-6},
 publisher            = {CEUR Workshop Proceedings},
}

@inproceedings{ngo2023csam,
  author    = {V. M. Ngo and C. Thorpe and S. McKeever},
  title     = {Analysing Child Sexual Abuse Activities in the Dark Web based on an Efficient CSAM Detection Algorithm},
  booktitle = {The 2nd Annual Trust and Safety Research Conference},
  address   = {Stanford, USA},
  month     = {September},
  year      = {2023}
}

@INPROCEEDINGS{Ngo.RIVF.2022,
  author={Ngo, Vuong M. and Thorpe, Christina and Dang, Cach N. and Mckeever, Susan},
  booktitle={2022 RIVF Int. Conf. on Computing and Communication Technologies (RIVF)}, 
  title={Investigation, Detection and Prevention of Online Child Sexual Abuse Materials: A Comprehensive Survey}, 
  year={2022},
  pages={707-713},
}

@article{Mazzarello-Springer:2022,
  title={Risk factors for sexual revictimization and dating violence in young adults with a history of child sexual abuse},
  author={Mazzarello, Olivia and Gagn{\'e}, Marie-Emma and Langevin, Rachel},
  journal={Journal of Child \& Adolescent Trauma},
  volume={15},
  number={4},
  pages={1113--1125},
  year={2022},
  publisher={Springer},
}

@article{Owusu-Addo-Elsevier:2023,
title = {Prevalence and determinants of sexual abuse among adolescent girls during the COVID-19 lockdown and school closures in Ghana: A mixed method study},
journal = {Child Abuse \& Neglect},
volume = {135},
pages = {105997},
year = {2023},
issn = {0145-2134},
author = {E. Owusu-Addo and others},
}

@article{Aguerri-Springer:2023,
  title={Old crimes reported in new bottles: the disclosure of child sexual abuse on Twitter through the case\# MeTooInceste},
  author={Aguerri, Jes{\'u}s C and Molnar, Lorena and Mir{\'o}-Llinares, Fernando},
  journal={Social Network Analysis and Mining},
  volume={13},
  number={1},
  pages={27},
  year={2023},
  publisher={Springer},
}

@article{dang2021hybrid,
  title={Hybrid deep learning models for sentiment analysis},
  author={Dang, Cach N and Moreno-Garc{\'\i}a, Mar{\'\i}a N and De la Prieta, Fernando},
  journal={Complexity},
  volume={2021},
  number={1},
  pages={9986920},
  year={2021},
  publisher={Wiley Online Library},
}

@article{yamashita2018convolutional,
  title={Convolutional neural networks: an overview and application in radiology},
  author={Yamashita, Rikiya and Nishio, Mizuho and Do, Richard Kinh Gian and Togashi, Kaori},
  journal={Insights into imaging},
  volume={9},
  pages={611--629},
  year={2018},
  publisher={Springer},
}

@article{palangi2016deep,
  title={Deep sentence embedding using long short-term memory networks: Analysis and application to information retrieval},
  author={Palangi, Hamid and others},
  journal={IEEE/ACM transactions on audio, speech, and language processing},
  volume={24},
  number={4},
  pages={694--707},
  year={2016},
  publisher={IEEE},
}

@article{Kissos-Elsevier:2020,
  title={Can artificial intelligence achieve human-level performance? A pilot study of childhood sexual abuse detection in self-figure drawings},
  author={Kissos, Limor and Goldner, Limor and Butman, Moshe and Eliyahu, Niv and Lev-Wiesel, Rachel},
  journal={Child Abuse \& Neglect},
  volume={109},
  pages={104755},
  year={2020},
  publisher={Elsevier},
}

@article{Gangwar-Elsevier:2021,
  title={AttM-CNN: Attention and metric learning based CNN for pornography, age and Child Sexual Abuse (CSA) Detection in images},
  author={Gangwar, Abhishek and Gonz{\'a}lez-Castro, V{\'\i}ctor and Alegre, Enrique and Fidalgo, Eduardo},
  journal={Neurocomputing},
  volume={445},
  pages={81--104},
  year={2021},
  publisher={Elsevier},
}

@inproceedings{Laranjeira-ACM:2022,
 author               = {Laranjeira, C. and Macedo, J. and Avila, S. and Santos, J.},
 year                 = {2022},
 title                = {Seeing without Looking: Analysis Pipeline for Child Sexual Abuse Datasets},
 booktitle            = {Proc. of 2022 ACM Conference on Fairness, Accountability, and Transparency (FAccT'22)},
 pages                = {2189–2205},
 publisher            = {ACM},
}

@article{Hole-Others:2022,
  title={Developing automated methods to detect and match face and voice biometrics in child sexual abuse videos},
  author={Hole, Martyn and Frank, Richard and Logos, Katie and Westlake, Bryce and Michalski, Dana and Bright, David and Afana, Erin and Brewer, Russell and Ross, Arun and Swearingen, Thomas and others},
  journal={Trends and Issues in Crime and Criminal Justice},
  number={648},
  pages={1--15},
  year={2022},
}

@article{Iftikhar.Wiley.2024,
    author = {Alam, Iftikhar and Basit, Abdul and Ziar, Riaz Ahmad},
    title = {Utilizing Age-Adaptive Deep Learning Approaches for Detecting Inappropriate Video Content},
    journal = {Human Behavior and Emerging Technologies},
    volume = {2024},
    number = {1},
    pages = {7004031},
    year = {2024}
}

@INPROCEEDINGS{Puentes-IEEE:2023,
  author={Puentes, Juanita and others},
  booktitle={2023 IEEE/CVF International Conference on Computer Vision Workshops (ICCVW)}, 
  title={Guarding the Guardians: Automated Analysis of Online Child Sexual Abuse}, 
  year={2023},
  volume={},
  number={},
  pages={3730-3734},
}

@inproceedings{Cook-ACM:2023,
    author = {Cook, Darren and Zilka, Miri and DeSandre, Heidi and Giles, Susan and Maskell, Simon},
    title = {Protecting Children from Online Exploitation: Can a Trained Model Detect Harmful Communication Strategies?},
    year = {2023},
    publisher = {ACM},
    booktitle = {Proceedings of the 2023 AAAI/ACM Conference on AI, Ethics, and Society},
    pages = {5–14},
    numpages = {10},
    series = {AIES '23}
}

@article{Chadaga-Springer:2023,
  title={An Explainable Framework to Predict Child Sexual Abuse Awareness in People Using Supervised Machine Learning Models},
  author={Chadaga, Krishnaraj and others},
  journal={Journal of Technology in Behavioral Science},
  pages={1--17},
  year={2023},
  publisher={Springer},
}

@article{Kaur.2024,
title = {Deep learning-based approaches for abusive content detection and classification for multi-class online user-generated data},
journal = {International Journal of Cognitive Computing in Engineering},
volume = {5},
pages = {104-122},
year = {2024},
issn = {2666-3074},
author = {Simrat Kaur and Sarbjeet Singh and Sakshi Kaushal},
}

@article{Mosa.2025,
  author    = {Mohamed Atef Mosa},
  title     = {Optimizing text classification accuracy: a hybrid strategy incorporating enhanced NSGA-II and XGBoost techniques for feature selection},
  journal   = {Progress in Artificial Intelligence},
  year      = {2025},
}

@article{Li.2025,
  author    = {Tingting Li and Ziming Zeng and Shouqiang Sun},
  title     = {A two-stage cyberbullying detection based on multi-view features and decision fusion strategy},
  journal   = {Applied Intelligence},
  volume    = {55},
  number    = {4},
  pages     = {294},
  year      = {2025},
}

@inproceedings{Akhter-IEEE:2020,
  title={Automatic detection of offensive language for urdu and roman urdu},
  author={Muhammad Pervez Akhter and Zheng Jiangbin and Irfan Raza Naqvi and Mohammed Abdelmajeed and Muhammad Tariq Sadiq},
  journal  ={IEEE Access},
  pages={91213 - 91226} ,
  year={2020},
  volume={8},
  organization={IEEE},
}

@article{NGO-CAN-2024,
title = {Discovering child sexual abuse material creators' behaviors and preferences on the dark web},
journal = {Child Abuse \& Neglect},
volume = {147},
pages = {106558},
year = {2024},
issn = {0145-2134},
author = {Vuong M. Ngo and Rahul Gajula and Christina Thorpe and Susan Mckeever},
}

\end{document}